\documentclass[sigconf]{acmart}

\usepackage{multirow}
\usepackage{multicol}
\usepackage{booktabs}
\usepackage{epsfig}

\usepackage{url}
\usepackage{array}
\usepackage{color}
\usepackage{subfigure}

\graphicspath{ {./images/} }

\copyrightyear{2020} 
\acmYear{2020} 
\setcopyright{acmcopyright}\acmConference[MM '20]{Proceedings of the 28th ACM International Conference on Multimedia}{October 12--16, 2020}{Seattle, WA, USA}
\acmBooktitle{Proceedings of the 28th ACM International Conference on Multimedia (MM '20), October 12--16, 2020, Seattle, WA, USA}
\acmPrice{15.00}
\acmDOI{10.1145/3394171.3416299}
\acmISBN{978-1-4503-7988-5/20/10}

\begin{document}

\title{
Towards Accurate Human Pose Estimation in Videos of Crowded Scenes
}
\author{Shuning Chang}
\authornote{Authors contributed equally to this work. Work done during internship at YITU Technology.}
\author{Li Yuan}
\authornotemark[1]
\email{yuanli@u.nus.edu}
\affiliation{
  \institution{National University of Singapore}
}

\author{Xuecheng Nie}
\affiliation{%
 \institution{YITU Technology}}

\author{Ziyuan Huang}
\affiliation{%
  \institution{National University of Singapore}
}

\author{Yichen Zhou}
\affiliation{%
  \institution{YITU Technology}
}

\author{Yunpeng Chen}
\affiliation{%
  \institution{YITU Technology}
}

\author{Jiashi Feng}
\affiliation{%
  \institution{National University of Singapore}}

\author{Shuicheng Yan}
\affiliation{%
  \institution{YITU Technology}}

\renewcommand{\shortauthors}{Chang and Yuan, et al.}
\begin{abstract}
  Video-based human pose estimation in crowded scenes is a challenging problem due to occlusion, motion blur, scale variation and viewpoint change, etc. Prior approaches always fail to deal with this problem because of (1) lacking of usage of temporal information; (2) lacking of training data in crowded scenes. In this paper, we focus on improving human pose estimation in videos of crowded scenes from the perspectives of exploiting temporal context and collecting new data. In particular, we first follow the top-down strategy to detect persons and perform single-person pose estimation for each frame. Then, we refine the frame-based pose estimation with temporal contexts deriving from the optical-flow. Specifically, for one frame, we forward the historical poses from the previous frames and backward the future poses from the subsequent frames to current frame, leading to stable and accurate human pose estimation in videos. In addition, we mine new data of similar scenes to HIE dataset from the Internet for improving the diversity of training set. In this way, our model achieves best performance on 7 out of 13 videos and 56.33 average w\_AP on test dataset of HIE challenge.
  
\end{abstract}


\maketitle

\section{Introduction}
Human pose estimation is important for many computer vision applications, including human action recognition, human-computer interaction, and video surveillance. Due to the viewpoint variance, appearance variance and cluttered background, pose estimation is a very challenging task for large scale image and video datasets. 
Recently, significant progress has been made in this area~\cite{sun2019HRNet}. However, the pose estimation in complex events~\cite{lin2020human} is still relatively new and a challenging problem. In this challenge of pose estimation on crowded scenes and complex events, we propose to obtain the pose of single images based pose estimation method, which can be applied to each video frame to get an initial pose estimation, and a further refinement through frames can be applied to make the pose estimation consistent and more accurate.

\begin{figure}[t]
\includegraphics[width=1.0\linewidth]{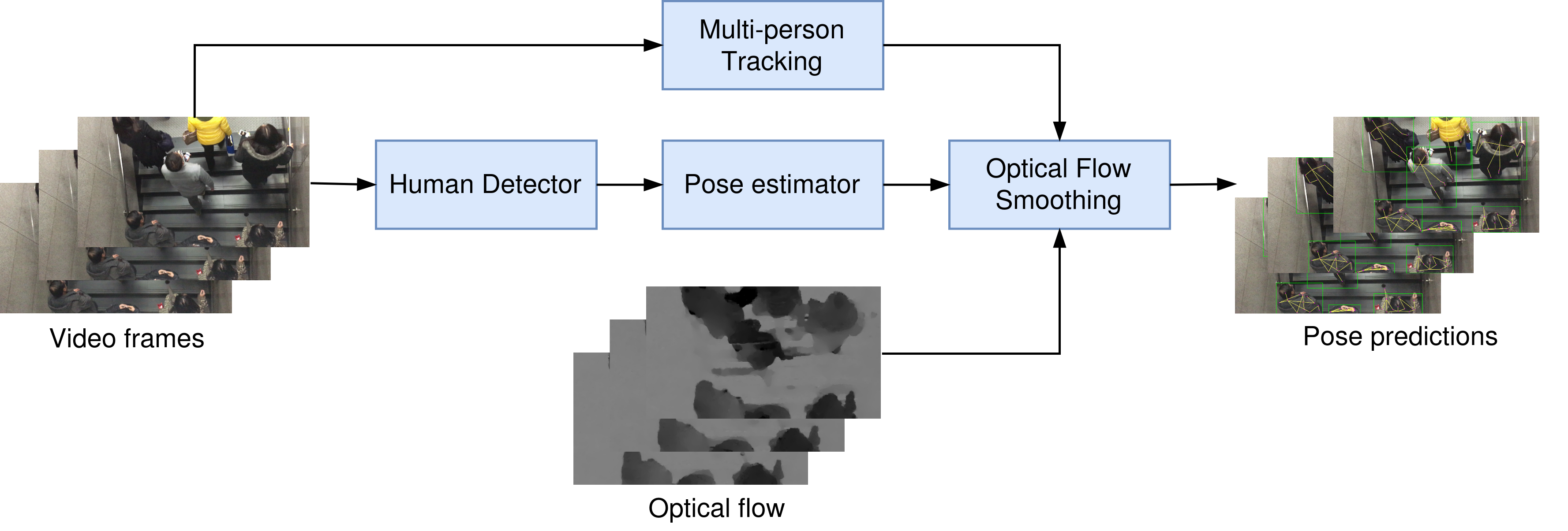}
\centering
\caption{The framework of our approach.}
\label{figure:structure}


\end{figure}

The general pipeline for the pose estimation method that we used can be divided into two parts, human detection and pose estimation, respectively. First, we use a human detection method in crowd scenes to detect the bounding box of highly-overlapping human instances in the detection phase. In the second step, we perform pose estimation on every box by two state-of-the-art single-person pose estimation models~\cite{xiao2018SimpleNet, sun2019HRNet}. During the pose estimation phase, we propose a optical flow smoothing algorithm to refine our pose predictions. The framework of our approach is shown in Figure \ref{figure:structure}. 

Since the problem is treated as a two-stage problem to be tackled one by one, each module will be introduced separately. The following of the report is organized as follows: Sec.~\ref{sec:human_det} investigates the common detection methods on HIE2020 challenge and also introduces our detailed method and experiments on human detection; Sec.~\ref{sec:pose_estimation} introduces the pose estimation as well as the final pose generation process. Sec.~\ref{sec:experiments} introduces the experiments and training details of pose estimation. Finally, Sec.~\ref{sec:conclusion} concludes the report.

\vspace{-2mm}

\section{Human Detection}
\label{sec:human_det}
The first step of human pose tracking is to detect the bounding boxes of person. As no validation set in HIE dataset~\cite{lin2020human}, we split the original training set as a new training set and validation set. Two splitting strategies are tried: splitting by image frames (5k for validation, 27k for training) and splitting by videos. We found that splitting by image will cause over-fitting and far away from the data distribution of the testing set. So we adopt the video-splitting strategy and split video 3,7,8 and 17 as the validation set and the rest videos as train data, in which 5.7k image frames for validation and the reset 27k images frames for training. Based on the train and validation set, we can conduct detection experiments on HIE. All the performance of models is tested by two metrics, Averaged Precision (AP) and MMR~\cite{dollar2011pedestrian}. AP reflects both the precision and recall ratios of the detection results; MMR is the log-average Miss Rate on False Positive Per Image (FPPI) in $[0.01,100]$, is commonly used in pedestrian detection. MR is very sensitive to false positives (FPs), especially FPs with high confidences will significantly harm the MMR ratio. Larger AP and smaller MMR indicates better performance.
\vspace{-3mm}

\subsection{Common Detection Frameworks}
There are mainly two different types of common detection frameworks: one-stage (unified) frameworks~\cite{redmon2016you, redmon2016yolo9000, liu2016ssd} and two-stage (region-based) framework~\cite{girshick2014rich, girshick2015fast, ren2015faster, he2017mask}. Since RCNN~\cite{girshick2014rich} has been proposed, the two-stage detection methods have been widely adopted or modified~\cite{ren2015faster, lin2017focal, lin2017FPN, cai2018cascade, zhou2018object, wang2019few, wang2019distilling}. Normally, the one-stage frameworks can run in real-time but with the cost of a drop in accuracy compared with two-stage frameworks, so we mainly adopt two-stage frameworks on HIE dataset. 

We first investigate the performance of different detection backbone and framework on HIE dataset, including backbone: ResNet152~\cite{he2016deep}, ResNeXt101~\cite{xie2017aggregated} and SeNet154~\cite{hu2018squeeze}, and different framework: Faster-RCNN~\cite{ren2015faster}, Cascade R-CNN~\cite{cai2018cascade}, and Feature-Pyramid Networks (FPN)~\cite{lin2017feature}. The experimental results on different backbone and methods are given in Table~\ref{tab:det_methods}. The baseline model is Faster RCNN with ResNet50, and we search hyper-parameters on the baseline model then apply to the larger backbone. From table~\ref{tab:det_methods}, we can find that the better backbone (ResNet152 and ResNeXt101) and combining advanced methods (Cascade and FPN) can improve the detection performance, but the SENet154 does not get better performance than ResNet152 even it has superior classification performance on ImageNet. So in our final detection solution, we only adopt ResNet152 and ResNeXt101 as the backbone.
\vspace{-2.5mm}

\subsection{Extra Data for Human Detection}
In the original train data, there are 764k person bounding boxes in 19 videos with 32.9k frames, and the testing set contains 13 videos with 15.1k frames. Considering the limited number of videos and duplicated image frames, the diversity of train data is not enough. And the train data and test data have many different scenes, thus extra data is crucial for training a superior detection model. Here we investigate the effects of different human detection dataset on HIE, including all the person images in COCO (COCO person, 64k images with 262k boxes)~\cite{lin2014microsoft}, CityPerson (2.9k image with 19k boxes)~\cite{zhang2017citypersons}, CrowndHuman (15k images with 339k boxes)~\cite{shao2018crowdhuman} and self-collected data (2k images with 30k boxes). We investigate the effects on different data based on Faster-RCNN with ResNet50 as the backbone. The experimental results are shown in Table~\ref{tab:det_data}. We can find that the CrowdHuman dataset achieves the largest improvement compared with other datasets, because the CrowdHuman is the most similar scenes with HIE, and both of the two datasets contain plenty of crowded scenes. COCO person contains two times of images than HIE train data, but merging the COCO person does not bring significant improvement and suffer more than three times train time, thus we only merge HIE with CrowdHuman and self-collected data to take a trade-off between detection performance and train time. 
\vspace{-3.5mm}

\subsection{Detection in Crowded Scenes}
As there are lots of crowded scenes in HIE2020 dataset, the highly-overlapped instances are hard to detect for the current detection framework. We apply a method aiming to predict instances in crowded scenes~\cite{chu2020detection}, named as ``CrowdDet''. The key idea of CrowdDet is to let each proposal predict a set of correlated instances rather than a single one as the previous detection method. The CrowdDet includes three main contributions for crowded-scenes detection: (1) an EMD loss to minimize the set distance between the two sets of proposals~\cite{tang2014detection}; (2). A refine module that takes the combination of predictions and the proposal feature as input, then performs a second round of predicting. (3). Set NMS, it will skip normal NMS suppression when two bounding boxes come from the same proposal, which has been proved works in crowded detection; We conduct experiments to test the three parts on HIE2020 dataset, and the results are shown in Table~\ref{tab:det_crowded}. Based on the results in the Table, we can find that the three parts do improve the performance in crowded detection. Meanwhile, we apply KD regularization~\cite{yuan2020revisiting} in the class's logits of the detection model, which can consistently improve the detection results by 0.5\%-1.4\%.

Finally, based on the above analysis, we train two detection models on HIE by combining extra data with the crowded detection framework: (1). ResNet152 + Cascade RCNN + extra data + emd loss + refine module + set NMS + KD regularization, whose AP is 83.21; (2). ResNeXt101 + Cascade RCNN + extra data + emd loss + refine module + set NMS + KD regularization, whose AP is 83.78;  Then two models are fused with weights 1:1.  
\vspace{-3.5mm}

\begin{table}[]
\begin{center}
\fontsize{8pt}{12pt}\selectfont
\caption{Performance comparison (AP and mMR) among different detection backbone and methods on HIE dataset. }
\begin{tabular}{l|c|c}
 \toprule
 Methods or Modules & AP (\%) & MMR (\%) \\
 \midrule
 Baseline (ResNet50 + Faster RCNN) & 61.68 & 74.01 \\
 ResNet152 + Faster RCNN &67.32 & 68.17\\
 ResNet152 + Faster RCNN + FPN &69.77 & 64.83\\
 SENet154 +  Faster RCNN + FPN &65.77 & 68.46\\
 ResNeXt101 + Faster RCNN + FPN &69.53 & 63.91\\
 ResNeXt101 + Cascade RCNN + FPN &71.32 & 61.58\\
 ResNet152 +  Cascade RCNN + FPN &71.06 & 62.55\\
 \bottomrule
\end{tabular}
\label{tab:det_methods}
\end{center}
\vspace{-3mm}
\end{table}

\begin{table}[]
\begin{center}
\fontsize{7pt}{12pt}\selectfont
\caption{The effects of using extra data for human detection on HIE dataset.}
\begin{tabular}{l|c|c}
 \toprule
 Validation set & AP (\%) & MMR (\%) \\
 \midrule
 HIE data & 61.68 & 74.01 \\
 HIE + COCO person &65.83 & 69.75\\
 HIE + CityPerson &63.71 & 67.43\\
 HIE + CrowdHuman &\textbf{78.22} & \textbf{58.33}\\
 HIE + self-collected data &\textbf{69.39} & \textbf{60.82}\\
 HIE + CrowdHuman + COCO + CityPerson &78.53 & 58.63\\
 \textbf{HIE + CrowdHuman + self-collected data}  &\textbf{81.03} & \textbf{55.58}\\
 HIE + all extra data &81.36 & 55.17\\
 \bottomrule
\end{tabular}
\label{tab:det_data}
\end{center}
\vspace{-3mm}
\end{table}

\begin{table}[]
\begin{center}
\fontsize{7pt}{12pt}\selectfont
\caption{Detection in Crowded Scenes on HIE dataset.}
\begin{tabular}{l|c|c}
 \toprule
 Validation set & AP (\%) & MMR (\%)\\
 \midrule
 ResNet50 + Faster RCNN + extra data & 81.36 & 55.17 \\
 + emd loss  &81.73 & 53.20\\
 + refine module &81.96 & 50.85\\
 + set NMS &\textbf{82.05} & \textbf{49.63}\\
 \bottomrule
\end{tabular}
\label{tab:det_crowded}
\end{center}
\vspace{-2mm}
\end{table}

\section{Pose Estimation}
\label{sec:pose_estimation}

In this section, we will introduce the networks we used to generate pose estimation and the optical flow smoothing algorithm serving for smoothing the pose predictions.

\subsection{Single-person Pose Estimators}

We adopt two state-of-the-art single-person pose estimation models, HRNet~\cite{sun2019HRNet} and SimpleNet~\cite{Xiao_2018_ECCV}, as our basic networks to generate pose predictions. Different from general high-to-low and low-to-high pattern, HRNet can maintain the high-resolution representations through the whole process and fuse multi-resolution representations simultaneously. SimpleNet is a simple and effective model, which just consists of a backbone network, ResNet in our work, declining the resolution of the feature map, and several deconvolutional layers producing the pose predictions. Additionally, for SimpleNet, we plug an FPN~\cite{lin2017FPN} structure in it to strengthen the performance of small person instances. Finally, we fuse the results of two models by averaging their heatmaps.

\subsection{Optical Flow Smoothing}

This task is based on videos, so the temporal information is a potentially available condition. Moreover, most people's actions in this dataset are not bouncing, just simple standing, sitting, and walking, so the poses from the same person are similar between adjacent frames. However, our single models cannot capture the temporal relationship. To solve this issue, we design an optical flow smoothing algorithm to smooth our pose predictions. 

We propose to smooth the current frame from the previous frame and the next frame by optical flow which is often expressed for temporal information. Given one human instance with joints coordinates set $\hat{J^{k}_i}$ in current frame $I^k$, first we compute $J^{k-1}_i$ in frame $I^{k-1}$ and the optical flow field $F_{k-1\xrightarrow{}k}$ between frame $I^{k-1}$ and $I^k$, then we can estimate the current frame joints coordinates set $\hat{J}^{k-1\xrightarrow{}k}_i$ in frame $I^k$ by propagating the joints coordinates set $J^{k-1}_i$ according to $F_{k-1\xrightarrow{}k}$. Specifically, for each joint location $(x, y)$ in $J^{k-1}_i$, the propagated joint location will be $(x + \delta x, y + \delta y$, where $\delta x$, $\delta y$ are the flow field values at joint location $(x, y)$. Similarly, we can estimate the current frame $\hat{J}^{k+1\xrightarrow{}k}_i$  from the next frame in the same way. Finally, we obtain the final predicted $J^{k}_i$ as follows:
\begin{equation}
\label{eqn1}
J^{k}_i = \alpha\cdot\hat{J}^{k-1\xrightarrow{}k}_i + \alpha\cdot\hat{J}^{k+1\xrightarrow{}k}_i + (1-2\alpha)\cdot \hat{J}^k_i,
\end{equation}
where the $\alpha$ is used to weighted sum the three terms.

The bottleneck of our method is how to track the same person in the adjacent frames. Some traditional work applies bounding box IoU (Intersection-over-Union) or pose similarity to link instances~\cite{Xiao_2018_ECCV}. However, there are numerous ultra crowded scenes in this dataset, which leads to severe occlusion and overlap, so the traditional methods would be problematic. Different previous work, we use a person Re-ID (person Re-identification) model to extract features to compute similarity. Compared with other methods, the Re-ID features focus on human appearance more, therefore, they are more suitable for this dataset. To verify our inference, we submit our tracking result to track 1 (private) server and achieve 61.0951\% on MOTA metric, which demonstrates the effect of our Re-ID features.

For the whole procedure of our optical flow smoothing algorithm, first, we utilize our person tracking model using Re-ID features to generate the person ID. Then, if the same IDs exist in the previous and next frames and their confidence scores are higher than a threshold, we will use Eqn \ref{eqn1} to smooth our pose estimation.



\begin{table} [t]
\fontsize{7pt}{10pt}\selectfont
\begin{center}
\caption{The top-3 results of HIE2020 testing set. The evaluation metric is w-AP(\%).To compare with the results for each video, we highlight the best results by red color and highlight the second one by blue color. Our approach achieves the best results on the vast majority of videos}
\label{result}
\begin{tabular}{c|c|c|c}
\toprule
Video Name & First Place & Ours & Third Place \\
\midrule
hm\_in\_waiting\_hall & \textcolor{blue}{64.5796} & \textcolor{red}{65.8270} & 58.5896\\
hm\_in\_bus & \textcolor{red}{56.0834} & \textcolor{blue}{55.4518} & 50.6453\\
hm\_in\_dining\_room2 & 22.2609 & \textcolor{red}{25.9449} & \textcolor{blue}{23.5640}\\
hm\_in\_lab2 & \textcolor{red}{72.7162} & \textcolor{blue}{70.3300} & 69.9452\\
hm\_in\_subway\_station & \textcolor{blue}{41.5776} & \textcolor{red}{47.2997} & 41.3249\\
hm\_in\_passage & \textcolor{blue}{88.9244} & \textcolor{red}{90.1478} & 86.5233\\
hm\_in\_fighting4 & \textcolor{blue}{57.1941} & \textcolor{red}{59.8902} & 56.3970\\
hm\_in\_shopping\_mall3 & \textcolor{blue}{61.2707} & \textcolor{red}{62.7075} & 60.9024\\
hm\_in\_restaurant & 58.2427 & \textcolor{red}{67.4902} & \textcolor{blue}{64.3151}\\
hm\_in\_accident & 53.1889 & \textcolor{red}{56.9365} & \textcolor{blue}{54.6401}\\
hm\_in\_stair3 & 47.7152 & \textcolor{blue}{49.1054} & \textcolor{red}{49.6768}\\
hm\_in\_crossroad & \textcolor{blue}{75.5781} & \textcolor{red}{75.7640} & 73.8597\\
hm\_in\_robbery & 51.2827 & \textcolor{red}{52.2467} & \textcolor{blue}{51.4776}\\
\midrule
Weighted Average & 57.5091 & 56.3375 & 55.1719\\
\bottomrule
\end{tabular}
\end{center}
\vspace{-3mm}
\end{table}

\begin{table*} 
\begin{center}
\fontsize{7pt}{10pt}\selectfont
\caption{Performance evaluation of different components in our method on the HIE testing set.}
\label{ablation}
\begin{tabular}{p{15mm}<{\centering}p{15mm}<{\centering}p{30mm}<{\centering}p{25mm}<{\centering}p{20mm}<{\centering}p{35mm}<{\centering}|c}
\toprule
HRNet & SimpleNet & Multi-scale Evaluation & Multi-scale Input & Extra Data & Optical Flow Smoothing & w-AP(\%) \\
\midrule
\checkmark & & & & & & 52.45 \\
\checkmark &  & \checkmark & & & & 52.90 \\
\checkmark & & & & \checkmark & & 53.82 \\
\checkmark & \checkmark & \checkmark & & \checkmark & & 55.52 \\
\checkmark & \checkmark & \checkmark & \checkmark & \checkmark & & 56.04 \\
\checkmark & \checkmark & \checkmark & \checkmark & \checkmark & \checkmark & 56.34 \\
\toprule
\end{tabular}
\end{center}
\vspace{-5mm}
\end{table*}

\section{Experiments}
\label{sec:experiments}
\subsection{Extra Data for Human Pose}

The original official training set contains about 660.5K annotated poses. Considering that large-scale similar data exist due to frame-wise annotation, it is necessary to collect extra data to improve the performance. The extra training data we used come from two aspects:
(1) We fuse three mainstreaming public pose estimation datasets, COCO, MPII, and AI Challenger, into our training data. The COCO dataset contains over 250k person instances labeled with 17 key points. The MPII dataset consists of 25K images including over 40K person instances with annotated 16 body joints. The AI Challenger dataset is composed of about 700K person instances with annotated 14 body joints. Since the annotated key points in these datasets are not totally overlapped with official labels, for each dataset we use respective overlapped key points for training.
(2) Self-collected data with similar scenes are merged into our training set. The number of poses is not over 30K, which is far less than the official training data.

All the extra data are randomly merged into the official training set. We do not explore more complicate data fusion strategy.
\vspace{-3mm}

\subsection{Training Details}
We extend the human detection box in height or width to a fix $4:3$ aspect ratio, and then crop the box from the image, which is resized to a fixed size, $256\times192$ or $384\times288$. The data augmentation includes random rotation$([-45^{\circ}, 45^{\circ}])$, random scale$(0.65, 1.35)$, and flipping. Half body data augmentation is also applied. The network of HRNet used is HRNet-W48 and the backbone of SimpleNet used is ResNet152. We implement our training using PyTorch. 
\vspace{-2.5mm}

\subsection{Testing Details}
We adopt a multi-scale evaluation during testing. Specifically, we rescale the detection box to obtain new bounding boxes with different scales, then crop them to the original size and flip them to acquire their flipped counterparts. The generated boxes are feed into the network to produce heatmaps. We average those heatmaps and search the highest response to obtain the locations of key points. The scale factors used are 0.7, 1.0, and 1.3. Moreover, it is easy to suffer redundancy and wrong boxes in the complex and crowded scenes. We apply Pose NMS~\cite{Fang_2017_ICCV} to eliminate similar and low-confidence redundancies. 
\vspace{-2.5mm}

\subsection{Results}

\begin{figure}[t!]
\begin{center}
\end{center}
\subfigure{
\includegraphics[scale=0.055]{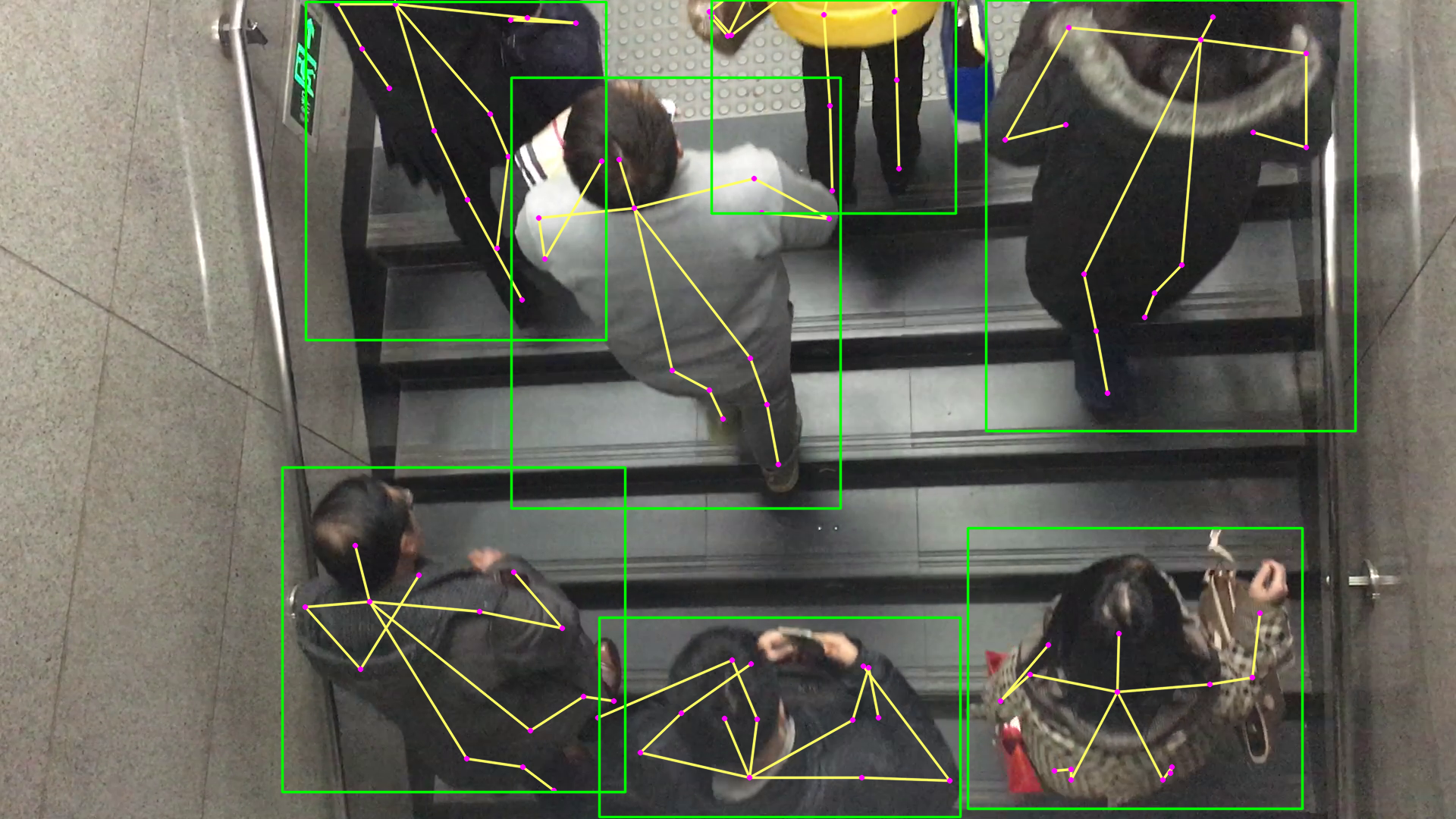}}
\subfigure{
\includegraphics[scale=0.093]{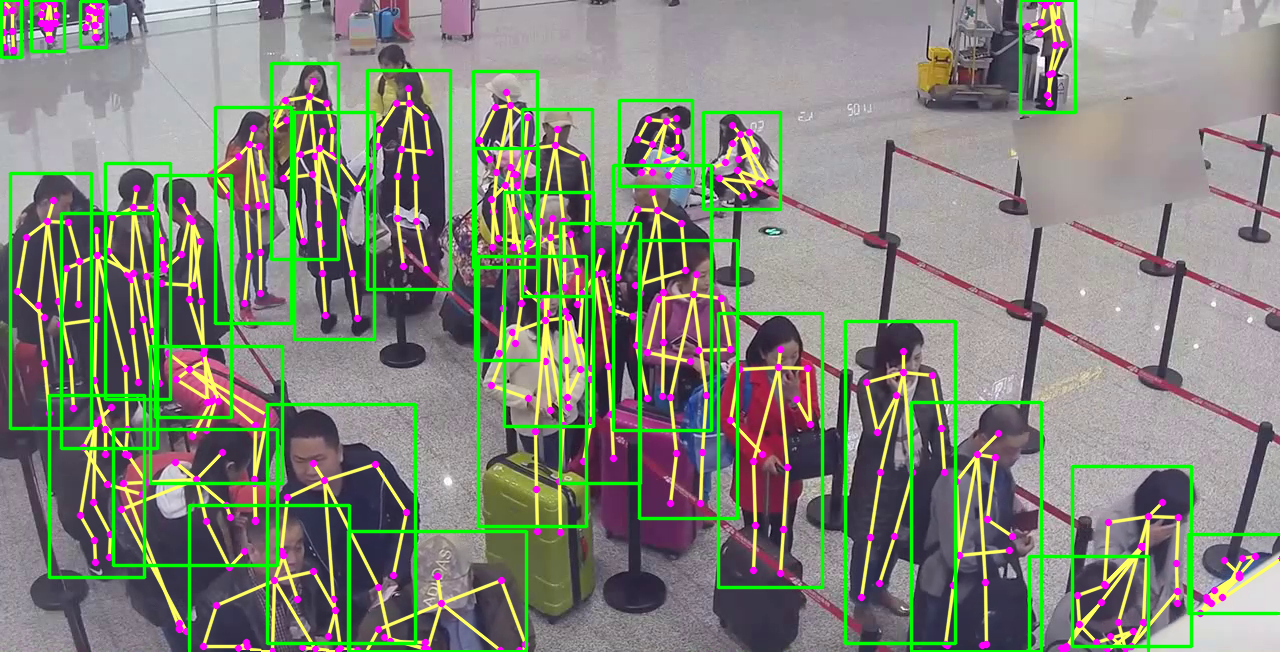}}
\subfigure{
\includegraphics[scale=0.1]{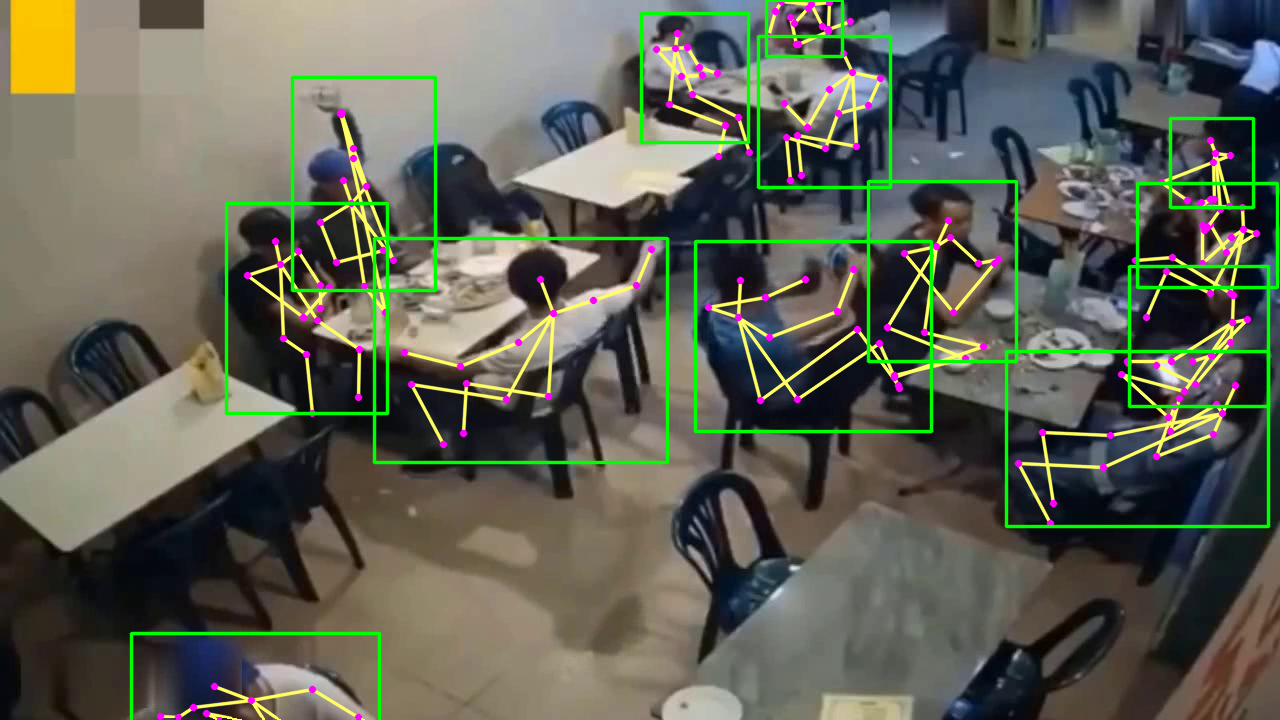}}
\subfigure{
\includegraphics[scale=0.15]{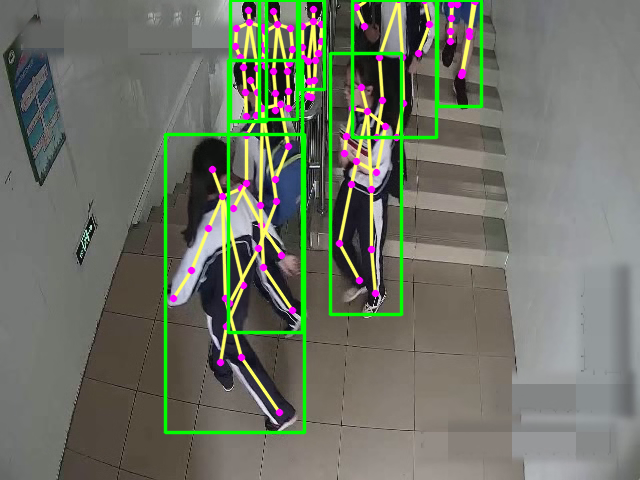}}
\caption{Example pose estimation results on the HIE2020 test set.}
\label{fig:visualization}
\end{figure}

The top-3 results of HIE2020 testing set are shown in Table \ref{result}. From our results for each video, we can see that our method achieves significant performance in the regular and high-resolution videos, such as "hm\_in\_passage" and "hm\_in\_crossroad". Our method performs poorly in the video with crowded scenes and low quality, \emph{e.g.}, we only get 25.94\% on the ultra crowded video "hm\_in\_dining\_room2", which is much lower than other videos. Our results for each video have remarkable performance. Even if compared with the first place, except "hm\_in\_bus" and "hm\_in\_lab2" are totally lower than them by 3\%, we achieve better performance in the rest videos. However, our weighted average result is 1.2\% lower than the winner. We analyse the possible reason is to exceed false positive predictions in our results. The false positive predictions are from two aspects: first, redundancy bounding boxes cause redundancy pose predictions; second, some small person instances are not involved in evaluation but we produce their poses.

We visualize some example of our pose estimation results in Figure \ref{fig:visualization}, which illustrates our approach can produce accurate pose predictions in the complex and crowded scenes.
\vspace{-3mm}

\subsection{Ablation Study}
In order to verify the performance of our components, we have done extensive experiments. The experiment results are shown in Table \ref{ablation}. Note that "Multi-scale Input" means training multiple groups of parameters by changing input size and fusing their results during testing. For each ablation experiment, if there is a \checkmark in the "Multi-scale Input" cell, the results is obtained by fusing input size $256\times 192$ and $384\times 288$; otherwise, the input size is just $256\times 192$. Extra data significantly boost our results by about 1.4\%, implying the effectiveness of large-scale data. Two fusion methods, model fusion between HRNet and SimpleNet and multi-scale input also improves our result tremendously by 1.7\% and 1.5\% 
respectively. Our post-processing algorithm, optical flow smoothing, can enhance the results by 0.3\%, which shows that it is effective.




\vspace{-3mm}
\section{Conclusion}
\label{sec:conclusion}
In this paper, we illustrate the approach we used in the HIE2020 Challenger pose estimation track. We adopt a top-down approach to address this complex and crowded scene issue. First, for human detection problems in crowded scenes, we add extra data to overcome the overfitting problem and apply one proposal for multiple predictions to relieve the difficulty of detecting highly-overlapping instances. Then, we apply our effective single-person pose estimation model to generate accurate pose predictions. To utilize temporal information, we design an optical flow smoothing algorithm to post-process our results.

\bibliographystyle{ACM-Reference-Format}
\bibliography{HIE}


\end{document}